\documentclass[letterpaper, 10 pt, conference]{ieeeconf}  

\IEEEoverridecommandlockouts                              

\usepackage{graphicx}%
\usepackage{multirow}
\usepackage{siunitx}
\usepackage{booktabs}
\usepackage{textcomp}
\usepackage{multirow}
\usepackage{comment}
\usepackage{dingbat}
\usepackage{hyperref} 
\usepackage{color,soul}
\usepackage[font=small, tableposition=top]{caption}
\title{\LARGE \bf
TEMPO-VINE: A Multi-Temporal Sensor Fusion Dataset for Localization and Mapping in Vineyards
}

\author{Mauro Martini$^{1}$, Marco Ambrosio$^{1}$, Judith Vilella-Cantos$^{2}$, Alessandro Navone$^{1}$, and Marcello Chiaberge$^{1}$
\thanks{$^{1}$ Department of Electronics and Communications, Politecnico di Torino, 10129, Torino, Italy.
        {\tt\small \{name.surname\}@polito.it}}%
\thanks{$^{2}$ University Institute for Engineering Research, Miguel Hernández University, Avda. de la Universidad s/n, Edificio Innova, Elche, 03202, Alicante, Spain.
        {\tt\small jvilella@umh.es}}%
\thanks{This work has been developed within the PoliTO Interdepartmental Centre for Service Robotics PIC4SeR.}
}

\let\oldtwocolumn\twocolumn
\renewcommand\twocolumn[1][]{%
    \oldtwocolumn[{#1}{
    \begin{center}
           \includegraphics[width=0.95\linewidth]{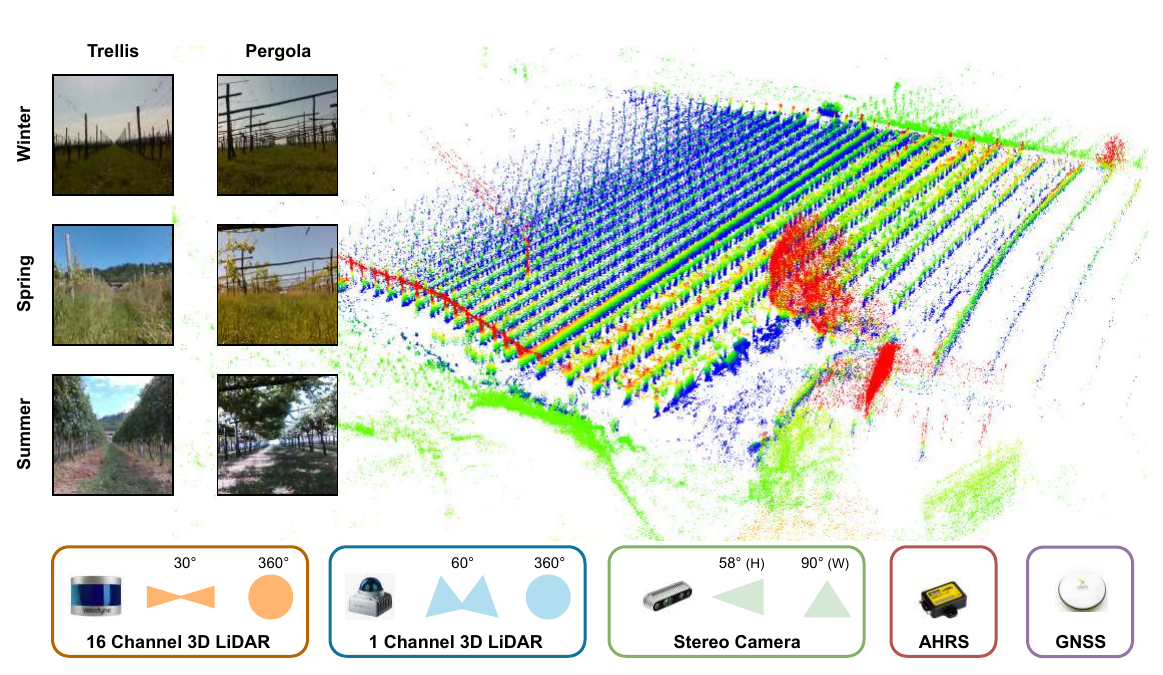}
           \captionof{figure}{TEMPO-VINE is a multi-seasonal dataset collected in trellis and pergola vineyards with different LiDARs, an RGB-D camera, an AHRS and a GNSS-RTK used for ground truth computation.}
           \label{fig:schema}
        \end{center}
    }]
}

\begin{document}

\maketitle
\thispagestyle{empty}
\pagestyle{empty}

\begin{abstract}
In recent years, precision agriculture has been introducing groundbreaking innovations in the field, with a strong focus on automation. However, research studies in robotics and autonomous navigation often rely on controlled simulations or isolated field trials. The absence of a realistic common benchmark represents a significant limitation for the diffusion of robust autonomous systems under real complex agricultural conditions. Vineyards pose significant challenges due to their dynamic nature, and they are increasingly drawing attention from both academic and industrial stakeholders interested in automation. 
In this context, we introduce the TEMPO-VINE dataset, a large-scale multi-temporal dataset specifically designed for evaluating sensor fusion, simultaneous localization and mapping (SLAM), and place recognition techniques within operational vineyard environments. TEMPO-VINE is the first multi-modal public dataset that brings together data from heterogeneous LiDARs of different price levels, AHRS, RTK-GPS, and cameras in real trellis and pergola vineyards, with multiple rows exceeding 100 m in length. In this work, we address a critical gap in the landscape of agricultural datasets by providing researchers with a comprehensive data collection and ground truth trajectories in different seasons, vegetation growth stages, terrain and weather conditions. The sequence paths with multiple runs and revisits will foster the development of sensor fusion, localization, mapping and place recognition solutions for agricultural fields. The dataset, the processing tools and the benchmarking results are available on the webpage\footnote{\url{https://sites.google.com/view/tempo-vine-dataset/dataset-overview}}.

\end{abstract}


\section{INTRODUCTION}

\begin{table*}[ht!]
\centering
\caption{Comparison with existing agricultural datasets.}
\label{dataset-comparison}
\begin{tabular}{@{}lllcccccc@{}}
\toprule
Dataset & Campaigns & Crop Type & Total length & 3D LiDAR & Camera & GPS & AHRS/IMU & Odom \\ \midrule
GREENBOT \cite{canadas2024multimodal} & 9 (2 months) & Tomato & 10.4 km & 2 & 1 stereo RGB & - & \checkmark & - \\
MAgro \cite{marzoa2024magro} & 9 (5 months) & Apple; Pear & 3.06 km & 1 & 2 stereo RGB & RTK & \checkmark & \checkmark \\
ARD-VO \cite{crocetti2023ard} & 11 (3 months) & Grapevine, olive. & ~28.8 km & 1 & \begin{tabular}[c]{@{}c@{}}2 stereo RGB, \\ 1 multispectral\end{tabular} & RTK & \checkmark & \checkmark \\
ROSARIO \cite{soncini2024rosario} & 6 (1 month) & Soybean & 7.3 km & - & stereo RGB-D & RTK & \checkmark & \checkmark \\
HORTO-3DLM \cite{barros2024spvsoap3d} & 6 (3 months) & \begin{tabular}[c]{@{}l@{}}Apple; Strawberry;\\ Cherry; Tomato\end{tabular} & 3.09 km & 1 & - & RTK & - & - \\
BLT \cite{polvara2024bacchus} & 11 (6 months) & \begin{tabular}[c]{@{}l@{}}Grapevine\\ (trellis)\end{tabular} & 0.5 km & 1 & 2 RGB-D & RTK & \checkmark & \checkmark \\
TEMPO-VINE (ours) & 13 (10 months) & \begin{tabular}[c]{@{}l@{}}Grapevine\\ (trellis and pergola)\end{tabular} & 38 km & 2 & 1 RGB-D & RTK & \checkmark & \checkmark \\ \bottomrule
\end{tabular}
\end{table*}

Precision agriculture is undergoing a technological transformation, with increasing integration of robotic systems and automation aimed at improving productivity, reducing labor costs, and enabling sustainable farming practices \cite{zhai2020decision}. Recent progress in autonomous navigation \cite{panda2023agronav, cerrato2024deep} and environmental perception \cite{ding2022recent} has demonstrated the potential of robotics in agricultural domain. Much of this research has focused on row-based crops, which represent roughly $75\%$ of planted fields in the United States \cite{Bigelow:263079}, addressing problems such as localization \cite{winterhalter2021localization}, path planning\cite{salvetti2023waypoint}, navigation\cite{man2020research}, harvesting \cite{hua2025harvesting}, pruning \cite{ NAVONE2025pruning}, and vegetation assessment \cite{feng2020yield}.
However, the development of reliable and generalizable solutions critically depends on the availability of high-quality datasets and standardized benchmarks to evaluate algorithmic performance in realistic settings.

The lack of datasets collected over entire seasons in operational agricultural environments that capture the complexities of the field represents a key limitation to research advancements. Robotic systems are often developed and validated in simulation \cite{martini2023enhancing} or isolated experiments, which fail to represent the variability in terrain, vegetation, and seasonal conditions. Among the most attractive crop fields worldwide, vineyards pose significant challenges for localization and mapping due to their dynamic appearance, structural variation, and changing environmental factors throughout the year.

To address this gap, we introduce TEMPO-VINE, a novel multi-temporal and multi-modal dataset designed for evaluating odometry, sensor fusion, simultaneous localization and mapping (SLAM), and place recognition methods in large-scale vineyard environments. TEMPO-VINE includes data from heterogeneous sensors, including two LiDARs (Velodyne VLP-16 and Livox Mid-360), a RGB-D camera, Attitude and Heading Reference System (AHRS), and RTK-GPS. 
Data have been collected through multiple campaigns from winter to autumn 2025, capturing diverse terrain, vegetation and weather conditions. The data are provided in a Robot Operating System (ROS) compatible format, with bags of data that can be easily played to test the desired algorithm and compute benchmarking metrics using the ground truth trajectories. Multiple runs for each campaign have been performed to augment the dataset with rich trajectory samples and to enable place recognition with revisited sequences.
TEMPO-VINE aims to foster the development of robust autonomous systems for agricultural robotics by providing a realistic, challenging, and diverse dataset for the robotics community.

The main contributions of this work are as follows:
\begin{itemize}
    \item The TEMPO-VINE dataset is a novel multi-seasonal and multi-modal benchmark for odometry, SLAM, sensor fusion, and place recognition in real-world vineyards covering both standard trellis and pergola vine architectures with row lengths exceeding $100$ meters.
    \item We present the first publicly available vineyards dataset that includes heterogeneous LiDAR data, supporting research in both high-performance and cost-effective robotic platforms.
    \item We provide ROS-compatible data format and ground-truth trajectories for repeated and loop-closure sequences, enabling reproducible evaluations.
    \item Through benchmarking experiments, we highlight the challenges posed by seasonal and structural variability in vineyards, emphasizing the need for more robust algorithms for perception and sensor fusion in agricultural fields.
\end{itemize}

\section{RELATED WORKS}
\subsection{Heterogeneous LiDAR datasets}
Light Detection and Ranging (LiDAR) sensors have proven highly suitable for place recognition tasks due to their invariance to illumination changes. Popular datasets, such as the Oxford \cite{maddern20171}, USyd \cite{zhou2020developing} and NCLT \cite{carlevaris2016university} datasets, collect data with this type of sensor during long-term campaigns to provide reliable data for developing and evaluating place recognition and simultaneous localization and mapping (SLAM) algorithms. These datasets capture LiDAR and vision information, along with seasonal changes, as they were recorded over the course of more than one year of consistent data collection sessions.

Recent studies have also focused on improving place recognition through deep learning, aiming at representations that remain robust under seasonal and appearance changes \cite{vilella2025minkunext}. Among the most common approaches is LiDAR–vision fusion, which combines the geometric invariance of LiDAR with the rich texture information provided by cameras \cite{komorowski2021minkloc++}. More recently, a growing trend is the fusion of different LiDAR sensors, since their varying fields of view (FoV) and coverage can produce higher-quality environmental maps \cite{jung2025helios}. Datasets such as HeliPR \cite{jung2024helipr} and GEODE \cite{chen2024heterogeneous} exemplify this direction, as they integrate data collected with multiple LiDAR sensors featuring different characteristics. These datasets also provide loop-closure trajectories for SLAM evaluation. Nevertheless, they lack seasonal variability, as they were not acquired over long-term campaigns, and the captured runs are relatively short and limited to urban environments.

\begin{figure}
    \centering
    \includegraphics[width=0.6\linewidth]{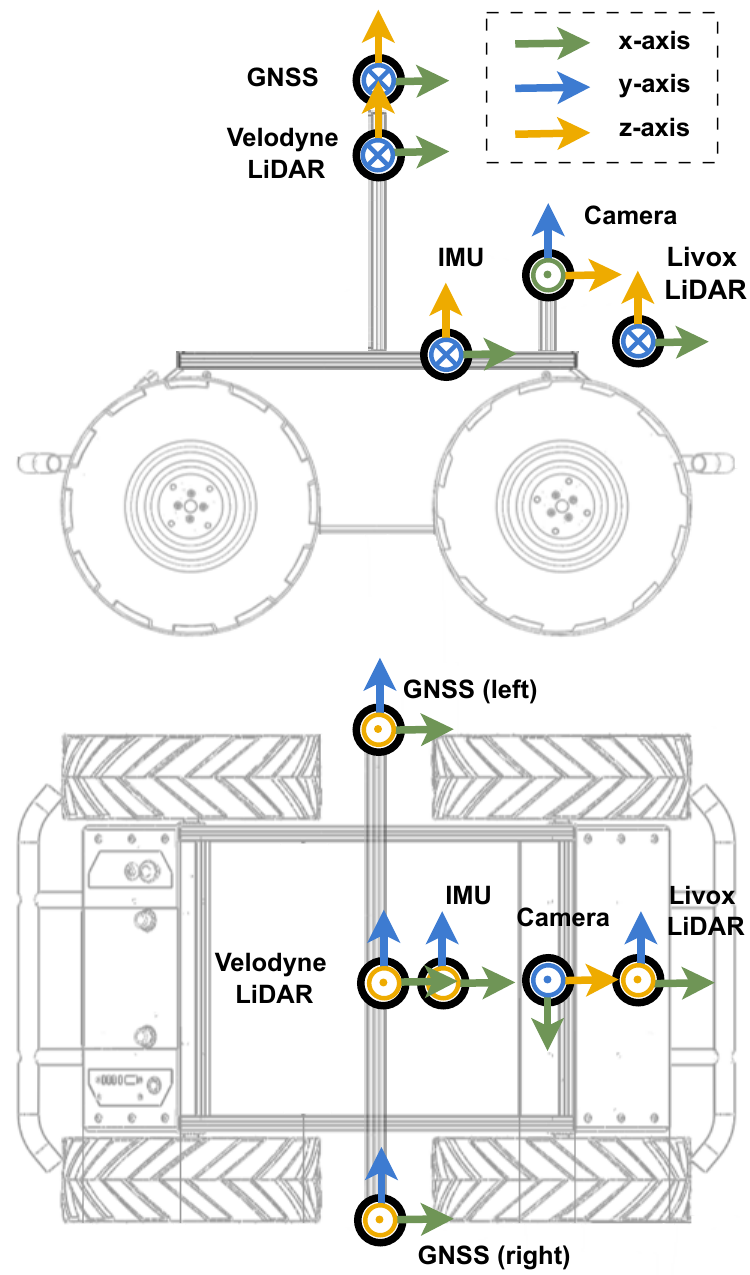}
    \caption{Schematic of the sensors disposition on the rover used for the data collection activity in the vineyards.}
    \label{fig:husky_config}
\end{figure}

\begin{table*}[ht]
\centering
\caption{Sensor Specifications}
\label{tab:sensors}
\begin{tabular}{@{}llc ccc c@{}} 
\toprule
Sensor & Type & Frequency & \multicolumn{3}{c}{FOV} & Resolution \\ 
\cmidrule(lr){4-6}
 &  &  & Range & Horizontal & Vertical &  \\ 
\midrule
3D LiDAR 1 & Livox Mid-360 & 10 Hz & 0.1--40 m & 360$^\circ$ & $-7^\circ$ to $+52^\circ$ & 0.2 m @1$\sigma$; $\pm$0.15$^\circ$ @1$\sigma$ \\
3D LiDAR 2 & \begin{tabular}[c]{@{}l@{}}Velodyne Puck\\ VLP-16\end{tabular} & 10 Hz & 100 m & 360$^\circ$ & $-15^\circ$ to $+15^\circ$ & 0.03 m; 0.1$^\circ$--0.4$^\circ$; 2.0$^\circ$ \\
RGB-D Camera & Intel Realsense D435 & 30 Hz & 0.3--3.0 m & 69$^\circ$ & 42$^\circ$ & 640$\times$480 px \\
AHRS-IMU & \begin{tabular}[c]{@{}l@{}}MicroStrain\\ 3DM-GX5\end{tabular} & 100 Hz & -- & -- & -- & 0.02 mg (accel); 0.003$^\circ$/s (gyro) \\
RTK-GPS & \begin{tabular}[c]{@{}l@{}}Swift Navigation\\ Duro\end{tabular} & 5 Hz & -- & -- & -- & $\sim$1--2 cm horiz.; $\sim$2--5 cm vert. \\
\bottomrule
\end{tabular}
\end{table*}

\subsection{Agricultural datasets}

While the previously mentioned urban datasets have become standard benchmarks for LiDAR-based perception and localization, agricultural environments pose unique challenges that require dedicated datasets. Unlike urban scenes with structured roads, buildings, and traffic rules, agricultural fields are highly dynamic and unstructured, with seasonal variations, repetitive patterns (e.g., vineyard rows), and heavy occlusions caused by vegetation.

Many agricultural datasets to date such as ARD-VO \cite{crocetti2023ard}, MAgro \cite{marzoa2024magro} or Rosario \cite{soncini2024rosario} primarily rely on RGB or multispectral imagery, which constrains their applicability to tasks beyond crop monitoring or visual detection. However, the lack of three-dimensional information limits progress on spatial perception problems such as mapping, navigation, and long-term place recognition. To address this gap, recent efforts like the Bacchus Long-Term (BLT) \cite{polvara2024bacchus} or the HORTO-3DLM \cite{barros2024spvsoap3d} datasets have started to incorporate 3D LiDAR recordings. Such developments open new opportunities for LiDAR-based place recognition (LPR) in agricultural environments. Nonetheless, those datasets only cover small fields with incomplete or costly sensor suites. 

Table \ref{dataset-comparison} compares agricultural datasets and shows that, to date, TEMPO-VINE offers the most complete sensor suite, making it a challenging and realistic benchmark for robot localization and mapping. Our dataset is the only one in the comparison that includes heterogeneous LiDAR; it has both a Velodyne VLP-16 and a Livox Mid-360. The Velodyne VLP-16 is a rotating LiDAR that generates a regular scan pattern with 16 simultaneous laser channels. In contrast, the Livox Mid-360 is a single-beam sensor that uses a non-repetitive scanning pattern to gradually cover its field of view. This results in different point distributions and sensing characteristics.

The TEMPO-VINE dataset was collected to address the need for agricultural datasets capturing long-term data throughout the crop's life cycle during different seasons. Vineyards are extremely dynamic environments due to vegetation growth, and, nonetheless, these factors should be combined with weather conditions and with the different field of view of the available sensors. Unlike previous vineyard-focused agricultural datasets, TEMPO-VINE provides multi-environment coverage, including both trellis and pergola vineyard structures, with rows exceeding $100$ m in length. The dataset also includes measurements from two complementary LiDAR sensors (one of them low-cost) together with additional modalities, enabling both intra-modal and cross-modal sensor fusion. Beyond geometry, it captures multi-seasonal variability, including different stages of crop development and ground cover conditions. Furthermore, the dataset is fully ROS-compatible, facilitating integration into robotic pipelines. 

\begin{figure*}
    \centering
    \includegraphics[width=1\linewidth]{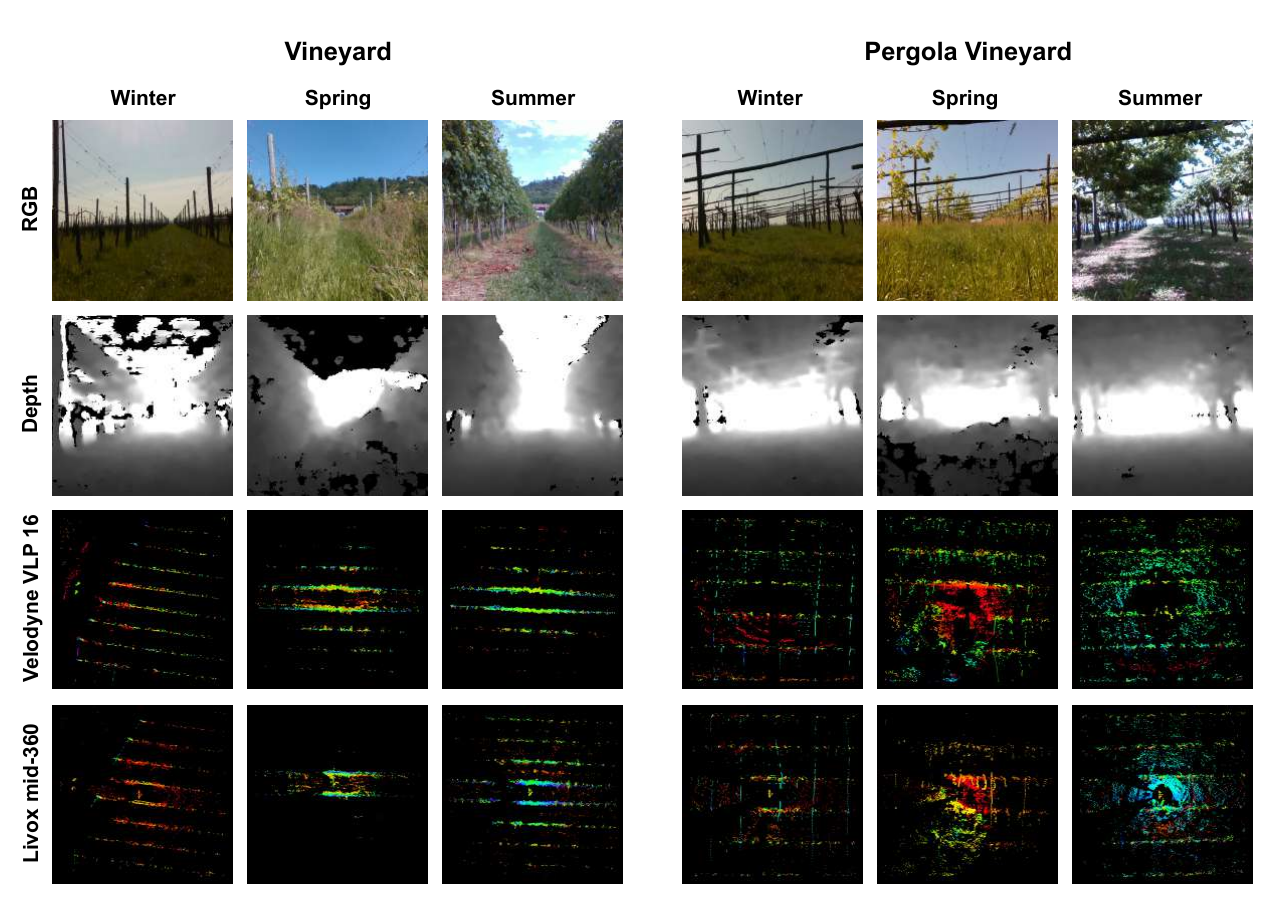}
    \caption{Samples of RGB-D images and LiDAR point cloud collected in the trellis and pergola vineyards in the same position over different seasons from winter to summer. Different vegetation growth stages represent a key dynamic environmental aspect for robotics navigation.}
    \label{fig:data_samples}
\end{figure*}

\section{Experimental Settings}

The recorded sensors have been selected to mix standard perception hardware adopted in robotics and low-cost sensing solutions. Indeed, beside the more advanced and costly Velodyne VLP-16 3D LiDAR, we also include a cheaper Livox Mid-360, a toroidal non-repetitive pattern 3D LiDAR that recently appeared on the market. The sensor suite also includes an RGB-D Intel Realsense D435 camera, providing aligned RGB and depth images, an accurate AHRS MicroStrain 3DM-GX5, and wheel encoders integrated in the rover's motors. Detailed sensor specifications are reported in Table \ref{tab:sensors}. 

The entire data collection has been performed with a Husky A200 rover from ClearPath Robotics \footnote{\href{https://share.google/8havUyAo1SyfeNKbJ}{Husky A200 rover datasheet}} (size 990 x 670 x 390 mm) equipped with multiple commercial sensors and an on-board computer with an Intel i7-6700E processor, 16 GB DDR4 RAM, and 1 TB SSD. Figure \ref{fig:husky_config} depicts the disposition of the different sensors on the robot for the data collection activity.

The rover has been manually teleoperated at a maximum linear velocity of $1 [m/s]$ and a maximum angular velocity of $2.0 [rad/s]$ to follow the desired trajectory. It has been remotely driven keeping it always in line of sight from the operator. Nonetheless, terrain irregularity and obstacles caused sudden drifts of the rover, which are difficult to promptly correct from long distances. This makes the trajectory collected highly realistic and comparable to an autonomous driving setting, with the rover not always staying at the centre of the rows.

\begin{table}[ht!]
\centering
\caption{Overview of field campaigns in the vineyards.}
\label{tab:campaign}
\resizebox{\columnwidth}{!}{%
\begin{tabular}{@{}llcccl@{}}
\toprule
\textbf{Campaign} & \textbf{Vineyard} & \textbf{Runs} & \textbf{Weather} & \textbf{Plant Growth} & \textbf{Grass Height} \\ \midrule
\textbf{ID 01} & Trellis & 01, 02     & Cloud        & Stem, no leaves             & Low ($\sim 5$ cm)    \\
12/02          & Pergola & 01, 02     & Cloud        & Stem, no leaves             & Low ($\sim 5$ cm)    \\ \midrule
\textbf{ID 02} & Trellis & 01, 02     & Sun          & Stem, no leaves             & Medium ($\sim 20$ cm)    \\
20/03          & Pergola & 01, 02     & Sun          & Stem, no leaves             & Medium ($\sim 20$ cm)    \\ \midrule
\textbf{ID 03} & Trellis & 01, 02     & Sun          & Stem, no leaves             & Tall ($>$40 cm)    \\
11/04          & Pergola & 01, 02     & Sun          & Stem, no leaves             & Tall ($>$40 cm)    \\ \midrule
\textbf{ID 04} & Trellis & 01, 02     & Sun          & Stem, few leaves            & Tall ($>$40 cm) \\
15/05          & Pergola & 01, 02     & Sun          & Stem, few leaves            & Tall ($>$40 cm) \\ \midrule
\textbf{ID 05} & Trellis & 01, 02     & Cloud        & Branches, leaves              & Medium ($\sim 20$ cm)   \\
06/06          & Pergola & 01, 02     & Cloud          & Branches, leaves              & Tall ($>$40 cm)   \\ \midrule
\textbf{ID 06} & Trellis & 01, 02     & Sun          & Branches, leaves              & Medium ($\sim 20$ cm)   \\
25/06          & Pergola & 01, 02     & Sun          & Branches, leaves              & Medium ($\sim 20$ cm)   \\ \midrule
\textbf{ID 07} & Trellis & 01, 02, 03 & Sun          & Branches, leaves              & Low ($\sim 5$ cm)    \\
11/07          & Pergola & 01, 02     & Sun          & Branches, leaves              & Medium ($\sim 20$ cm)    \\ \midrule
\textbf{ID 08} & Trellis & 01, 02, 03 & Sun          & Branches, leaves, fruit       & Low ($\sim 5$ cm)    \\
28/07          & Pergola & 01, 02     & Sun          & Branches, leaves, fruit       & Low ($\sim 5$ cm)    \\ \midrule
\textbf{ID 09} & Trellis & 01, 02, 03 & Sun          & Branches, leaves, fruit       & Medium ($\sim 20$ cm) \\
14/08          & Pergola & 01, 02     & Sun          & Branches, leaves, fruit       & Low ($\sim 5$ cm) \\ \midrule
\textbf{ID 10} & Trellis & 01, 02, 03 & Cloud          & Branches, leaves, fruit       & Medium ($\sim 20$ cm) \\
11/09          & Pergola & 01, 02     & Cloud        & Branches, leaves, fruit       & Tall ($>$40 cm) \\ 
\midrule
\textbf{ID 11} & Trellis & 01, 02, 03 & Sun              & Branches, leaves, fruit             & Low ($\sim 5$ cm)   \\
02/10          & Pergola & 01, 02     & Sun              & Branches, leaves, fruit             & Low ($\sim 5$ cm)    \\
\midrule
\textbf{ID 12} & Trellis & 01, 02, 03 & Cloudy               & Branches, leaves             & Low ($\sim 5$ cm)    \\
30/10          & Pergola & 01, 02     & Cloudy               & Branches, leaves             & Low ($\sim 5$ cm)    \\ \midrule
\textbf{ID 13} & Trellis & 01, 02, 03 & Cloudy              & Branches, leaves            & Medium ($\sim 20$ cm)   \\
19/11          & Pergola & 01, 02     & Cloudy              & Branches, leaves            & Medium ($\sim 20$ cm)   \\ 
\bottomrule
\end{tabular}%
}
\end{table}

\begin{figure}[h!]
     \centering
     \includegraphics[width=\linewidth]{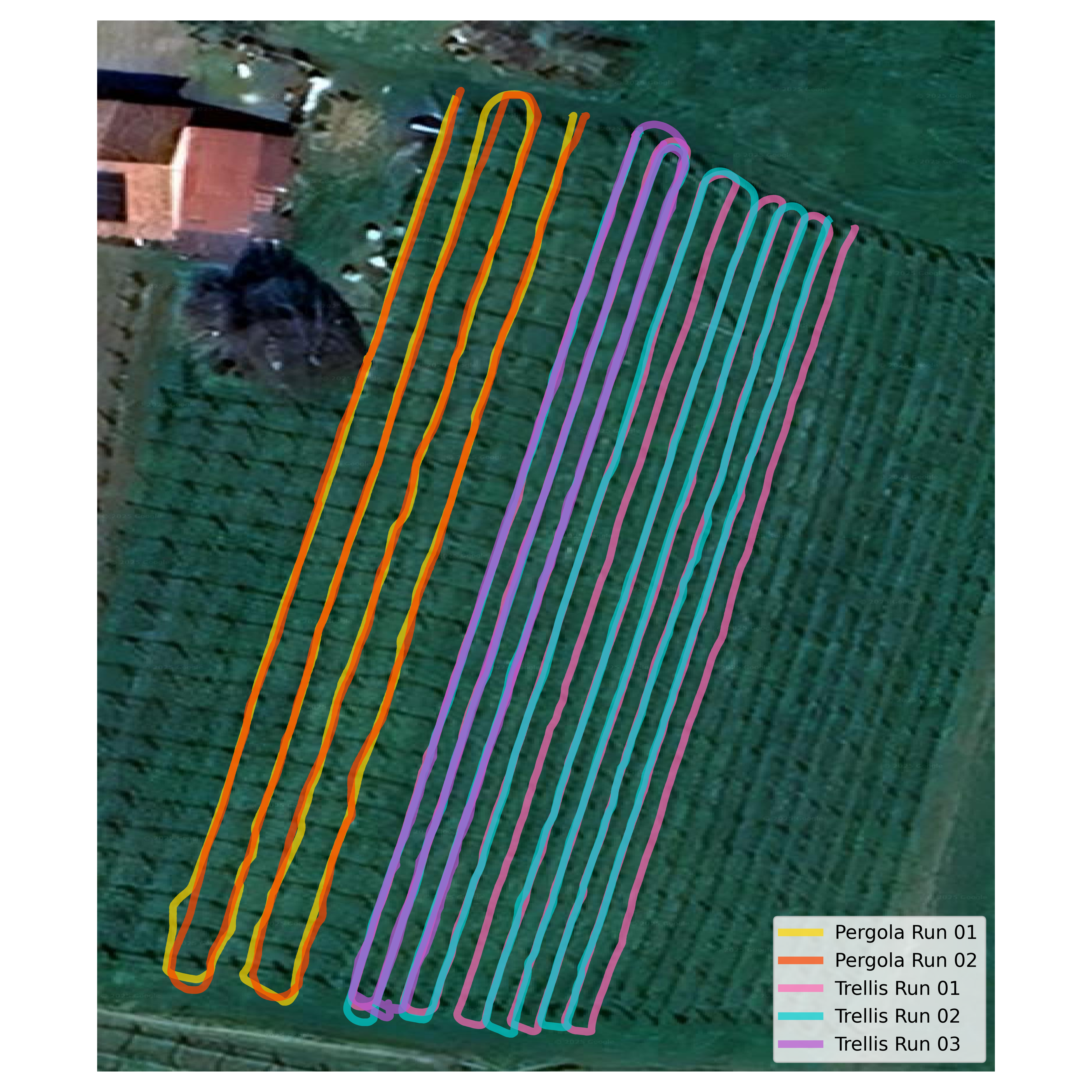}
      \caption{Trajectories for the three runs conducted in the trellis vineyard and in the pergola vineyards, overlaid with the satellite image of the field.}
     \label{fig:satellite_maps}
 \end{figure}

  \begin{figure}[h!]
     \centering
     \includegraphics[width=\linewidth]{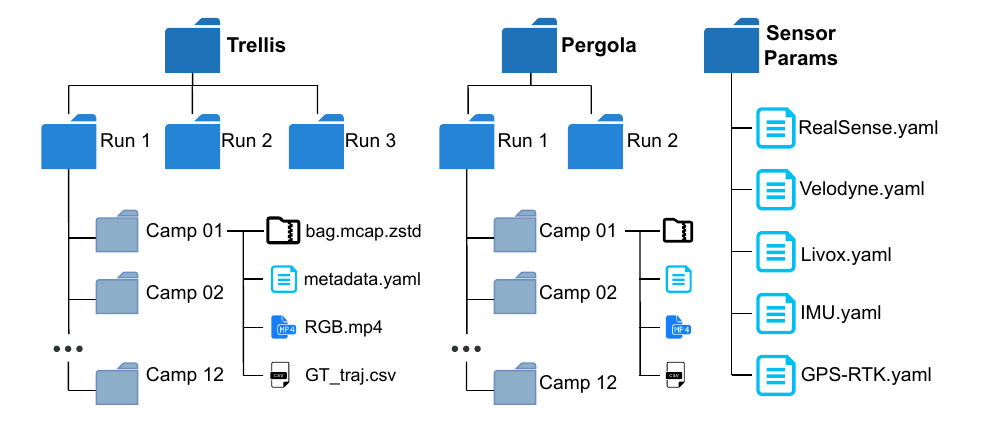}
     \caption{File structure of the TEMPO-VINE dataset: the folders are organized to easily select the vineyard field, the run and the campaign date. For each experiment a bag file with metadata containing all the complete sensor data stream is provided, together with the RGB camera video and the ground truth trajectory file.}
     \label{fig:file_structure}
 \end{figure}
 
\section{Description of the Dataset}
In this section, the dataset composition is presented, starting from the description of the data collection campaigns and trajectories performed on the field. Then, the ground truth trajectory computation is explained, together with the file structure and organization of the dataset.

\subsection{Vineyards Campaigns and Trajectories}
The TEMPO-VINE dataset currently comprises sensor data collected over 10 experimental campaigns distributed from February to September 2025 (with two other campaigns already scheduled in October and November that will be added upon approval in the review stage).

In each campaign multiple runs have been conducted in two different vineyards: 
\begin{itemize}
    \item A trellis Guyot vineyard with size $110 \times 25$ [m]. The data have been recorded using 10 rows of the field. Each row has an average width of 2 [m].
    \item A pergola vineyards with 4 rows and size $110 \times 15$ [m]. Each row has an average width of 3.5 [m].
\end{itemize}
Both vineyards are located in Agliè, Torino, Italy. The peculiarity of the dataset is its multi-seasonal temporal span, offering a wide range of diverse field conditions. Table \ref{tab:campaign} reports the entire sequence of campaigns conducted, the number of runs and details about the state of the vineyard.
The growth of plants' foliage over seasons, the presence of grass at different heights, and the weather conditions significantly shape the landscape of the vineyards over both long and short time periods, making it a highly dynamic and challenging environment for navigation. These factors influence the distribution of data taken from sensors, generating strong domain gaps in both images and point clouds and hindering the performance of machine learning, localization and place recognition algorithms. Some examples of data samples from the two LiDARs and the RGB-D camera collected in different seasons on the two vineyards are shown in Fig. \ref{fig:data_samples}. As can be seen in the selected samples, point clouds registered in the same position at different times drastically change, as well as the visual appearance of the row. Moreover, the two LiDARs offer different perspectives of the same row due to their specific field of view and position on the rover. This is extremely clear during Spring samples reported in Fig. \ref{fig:data_samples}.
 
The runs performed follow a specific pattern among the rows of the two vineyards to guarantee consistency and reproducibility of the algorithms' performance over different seasonal conditions, and to enable place recognition with multiple revisits of the same rows.
The complete visualization of the different trajectories performed in each campaign is visually shown in Fig. \ref{fig:satellite_maps}, overlaid on the satellite map of the entire vineyard. In the classic trellis vineyards (Fig. \ref{fig:satellite_maps}, top image), Run 01 is carried out traversing 10 rows continuously, while Run 02 covers the same path for the first 4 rows, then the other 4 rows are traversed in the opposite direction. Run 03 has been collected starting from the summer months when the vegetation reached a significant thickness, and it is looping over the first 3 rows to support the place recognition task in the most challenging scenario. In the pergola (Fig. \ref{fig:satellite_maps}, bottom image), we performed 2 runs with full coverage trajectories in the 4 available rows in all the months.

\subsection{Dataset Organization and Format}
The file structure of the TEMPO-VINE dataset is schematically illustrated in Figure \ref{fig:file_structure}. 
The structure is thought to easy access data of a specific experiment, choosing the vineyard field first, then the run, and finally the campaign. Each campaign folder contains 4 files: the compressed bag file \verb|bag.mcap.zstd| with all the raw stream data from all the sensors, the \verb|metadata.yaml| with all the bag information, the complete \verb|RGB.mp4| video recording from the on-board camera, and the ground truth trajectory file \verb|GT_traj.csv|. The bag file is ROS compatible, but does not necessarily require ROS installation to be played.
Sensors settings are a common feature for all the campaigns, hence the parameter files of each sensors are contained in the separate \textit{Sensor Params} folder in a \verb|.yaml| format. 

\begin{table}[]
\caption{Results of SLAM algorithms for each sensor tested in winter (March) and summer (July) in both the trellis (Run 02) and pergola (Run 01) vineyards.}
\label{tab:slam-eval}
\begin{tabular}{@{}ll|rrrr@{}}
\toprule
\multirow{3}{*}{\textbf{Sensor}} & \multirow{3}{*}{\textbf{RMSE}} & \multicolumn{2}{c}{\textbf{Trellis}} & \multicolumn{2}{c}{\textbf{Pergola}} \\ \cmidrule(l){3-6} 
 &  & \multirow{2}{*}{\textbf{02-R02}} & \multicolumn{1}{r|}{\multirow{2}{*}{\textbf{07-R02}}} & \multirow{2}{*}{\textbf{02-R01}} & \multirow{2}{*}{\textbf{07-R01}} \\
 &  &  & \multicolumn{1}{r|}{} &  &  \\ \midrule
\multirow{3}{*}{\begin{tabular}[c]{@{}l@{}}RGBD Camera\\ (RTAB-Map \cite{rtabmap})\end{tabular}} & ATE$_t$ & 8.32 & \multicolumn{1}{r|}{5.97} & 4.58 & 1.72 \\
 & RPE$_r$ & 0.03 & \multicolumn{1}{r|}{0.03} & 0.09 & 0.02 \\
 & RPE$_t$ & 0.21 & \multicolumn{1}{r|}{0.26} & 2.21 & 0.24 \\ \midrule
\multirow{3}{*}{\begin{tabular}[c]{@{}l@{}}Velodyne VLP-16\\ (Fast-LIO \cite{fastlio})\end{tabular}} & ATE$_t$ & 23.93 & \multicolumn{1}{r|}{26.46} & 0.58 & 1.54 \\
 & RPE$_r$ & 0.04 & \multicolumn{1}{r|}{0.05} & 0.06 & 0.05 \\
 & RPE$_t$ & 0.25 & \multicolumn{1}{r|}{0.33} & 0.36 & 0.24 \\ \midrule
\multirow{3}{*}{\begin{tabular}[c]{@{}l@{}}Velodyne VLP-16 \\ (LIO-SAM \cite{liosam})\end{tabular}} & ATE$_t$ & 1.93 & \multicolumn{1}{r|}{3.30} & 3.78 & 1.09 \\
 & RPE$_r$ & 0.45 & \multicolumn{1}{r|}{0.57} & 0.94 & 0.74 \\
 & RPE$_t$ & 0.03 & \multicolumn{1}{r|}{0.03} & 0.04 & 0.03 \\ \midrule
\multirow{3}{*}{\begin{tabular}[c]{@{}l@{}}Livox MID-360\\ (Fast-LIO \cite{fastlio})\end{tabular}} & ATE$_t$ & 1.37 & \multicolumn{1}{r|}{23.63} & 2.94 & 2.20 \\
 & RPE$_r$ & 0.04 & \multicolumn{1}{r|}{0.05} & 0.06 & 0.03 \\
 & RPE$_t$ & 0.17 & \multicolumn{1}{r|}{0.38} & 0.32 & 0.22 \\ \bottomrule
\end{tabular}
\end{table}

\begin{figure}
     \centering
     \includegraphics[width=\linewidth]{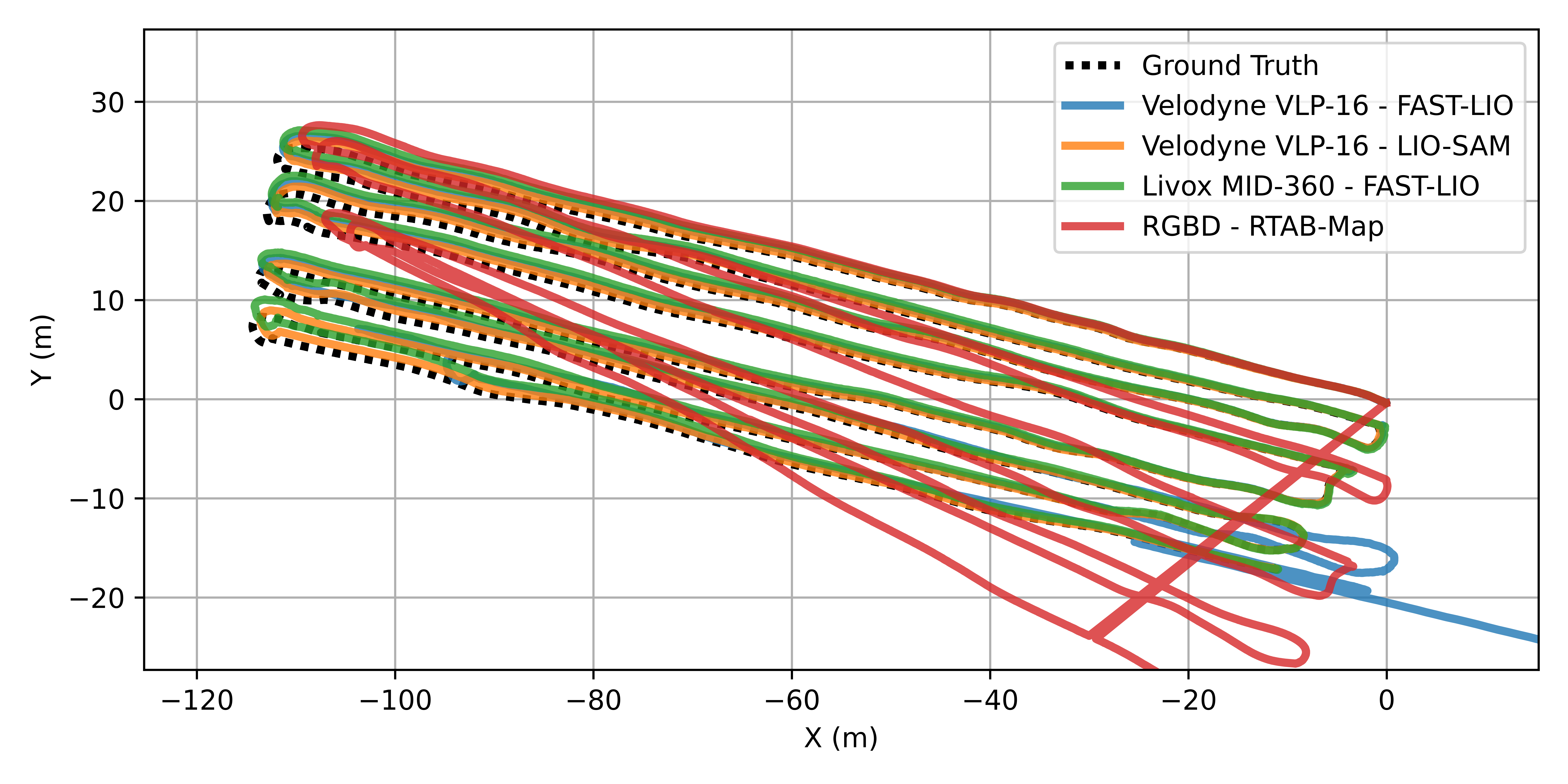}
     \includegraphics[width=\linewidth]{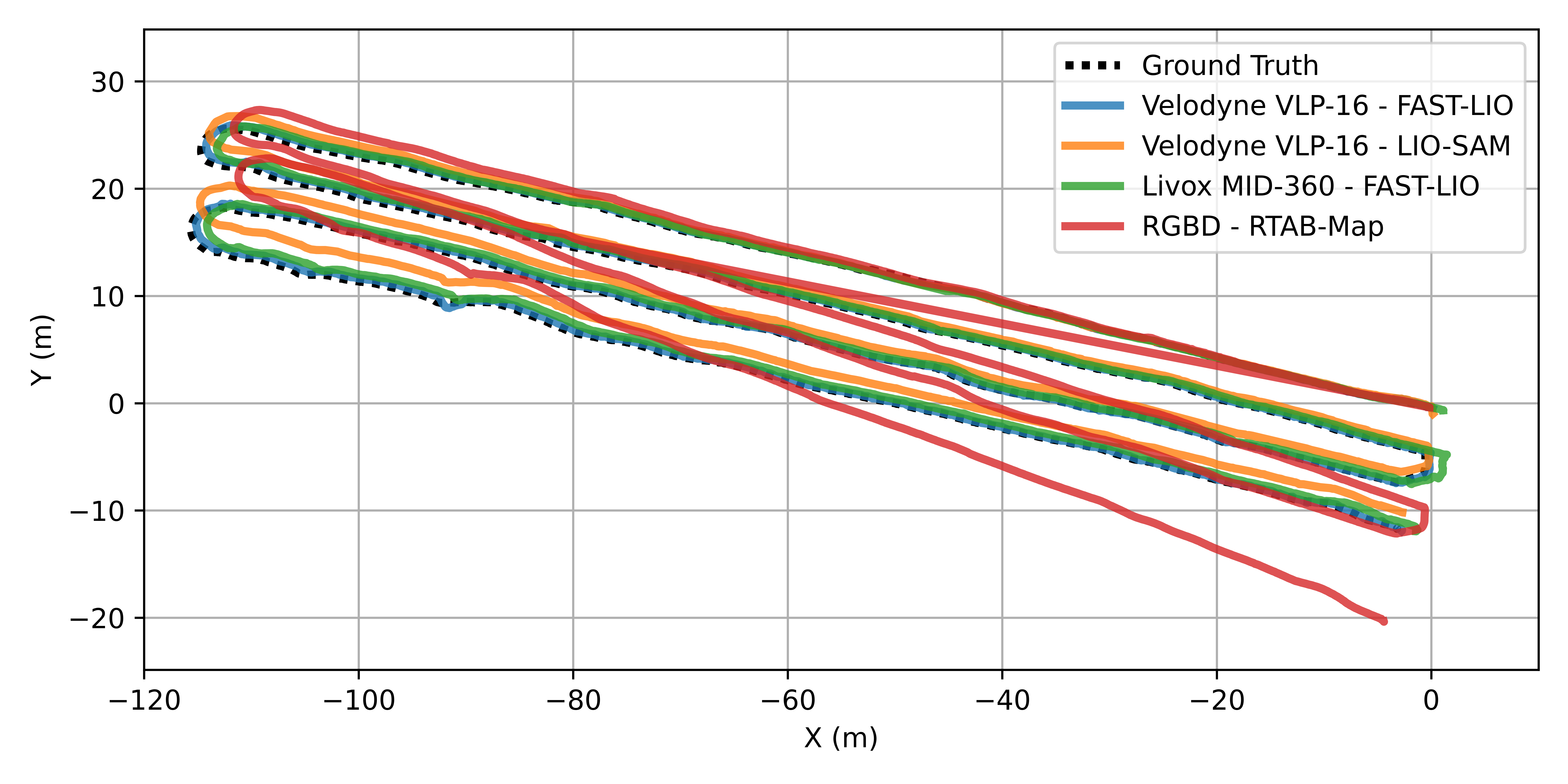}
      \caption{SLAM results and ground truth trajectories on trellis (top) and pergola (bottom) vineyards from campaign 02 (March) with two LiDARs and the RGB-D camera.}
     \label{fig:slam_results}
 \end{figure}
 
\subsection{Ground Truth} 
Ground truth is obtained using a high-precision GNSS-RTK receiver in combination with a high-end AHRS. Data from both sources are integrated to estimate the robot center’s position with a temporal resolution of 0.2 seconds. For each sensor, the extrinsic transformation to the robot center is specified. The ground truth trajectory is represented in a text file, where each line contains the timestamp, translation, and rotation (expressed as a quaternion) in a local ENU frame. A geo-referenced trajectory is also provided in the same format, with translation replaced by latitude, longitude, and altitude. This format, referred to as TUM, was selected to ensure compatibility with existing evaluation tools.

\section{DATASET EVALUATION}
In this section, we validate the data testing state-of-the-art algorithms for SLAM, both LiDAR and visual SLAM, and place recognition in the vineyard scenario. The results demonstrate the performance obtained using different sensors and the complexity generated from vegetation conditions in winter and summer.

\subsection{SLAM results} 
To assess the SLAM performance of the proposed dataset, three state-of-the-art ROS2-compatible algorithms were evaluated: RTAB-Map \cite{rtabmap} for RGB-D data, Fast-LIO \cite{fastlio} and LIO-SAM \cite{liosam} for LiDAR data. Experiments were conducted using data from campaigns 02 (March) and 07 (July), covering different seasonal conditions and two vineyard training systems: trellis and pergola. Each sensor was tested in each condition, resulting in 12 experiments in total.
Performance was evaluated against ground truth using Absolute Pose Error (APE, in meters) and Relative Pose Error (RPE), with RPE$_t$ representing translational error (in meters) and RPE$_r$ rotational error (unitless). APE values exceeding 4 meters were considered failures, indicating SLAM divergence.
LIO-SAM obtained the most robust result overall, performing successfully in all seasons. However, the algorithm is compatible only with ring-organized data like Velodyne. Fast-LIO with Livox data achieved precise performace, only failing in the summer trellis scenario. With Velodyne, Fast-LIO performed well in the pergola but failed in the final rows of the trellis. In contrast, RTAB-Map consistently underperformed across all conditions, highlighting the challenges posed by the environment for RGB-D-based SLAM.
Results are summarized in Table~\ref{tab:slam-eval} and Figure~\ref{fig:slam_results} shows the trajectories of the experiments from campaign 02.

\begin{table}
    \centering
    \caption{Place recognition quantitative analysis with Scan Context\cite{kim2018scan}: Recall@5 and Precision@5 for different thresholds and seasons in trellis vineyard (Run 01).}
    \begin{tabular}{llccc|ccc}
    \toprule
    \multirow{2}{*}{\textbf{Sensor}} &
    \multirow{2}{*}{\textbf{Camp}} & \multicolumn{3}{c}{\textbf{Recall}} & \multicolumn{3}{c}{\textbf{Precision}} \\ 
               &    & 10m & 15m & 20m              & 10m & 15m & 20m\\
        \midrule
        \multirow{3}{*}{VLP-16}
         & 02-March & 0.98 & 0.99 & 0.99 & 0.87 & 0.91 & 0.94\\
         & 04-May & 0.85 & 0.90 & 0.93 & 0.58 & 0.655 & 0.71 \\
         & 06-June & 0.27 & 0.41 & 0.53 & 0.10 & 0.19 & 0.29 \\
         \midrule
         \multirow{3}{*}{Mid-360}
         & 02-March & 0.76 & 0.83 & 0.87 & 0.38 & 0.49 & 0.59\\
         & 04-May & 0.41 & 0.55 & 0.68 & 0.17 & 0.26 & 0.36\\
         & 06-June & 0.28 & 0.44 & 0.55 & 0.11 & 0.20 & 0.31 \\
         \bottomrule
    \end{tabular}
    \label{pr-results-table-sc}
\end{table}

\begin{table}
    \centering
    \caption{Place recognition quantitative analysis of learned methods: Recall@1\% and Recall@1.}
    \begin{tabular}{cc|c|cc}
    \toprule
    \textbf{Sensor} &
    \textbf{Camp} & \textbf{Backbone} & \textbf{Recall@1\%} & \textbf{Recall@1} \\ 
        \midrule
        \multirow{3}{*}{VLP-16}
         & 02-March & PointNetVLAD & 0.68 & - \\
         & 02-March & MinkLoc3Dvs & 0.42 & 0.32 \\
         & 04-May & PointNetVLAD & 0.52 & -\\
         & 04-May & MinkLoc3Dv2 & 0.49 & 0.30\\
         & 06-June & PointNetVLAD &  0.47 & -\\
         & 06-June & MinkLoc3Dv2 &  0.38 & 0.30\\
         \midrule
         \multirow{3}{*}{Mid-360}
         & 02-March & PointNetVLAD & 0.59 & - \\
         & 02-March & MinkLoc3Dv2 & 0.39 & 0.32 \\
         & 04-May & PointNetVLAD & 0.46 & -\\
         & 04-May & MinkLoc3Dv2 & 0.35 & 0.30\\
         & 06-June & PointNetVLAD &  0.50 & -\\
         & 06-June & MinkLoc3Dv2 &  0.40 & 0.29\\
         \bottomrule
    \end{tabular}
    \label{pr-results-table-dl}
\end{table}

\begin{figure}[ht]
    \centering
    \includegraphics[width=\linewidth]{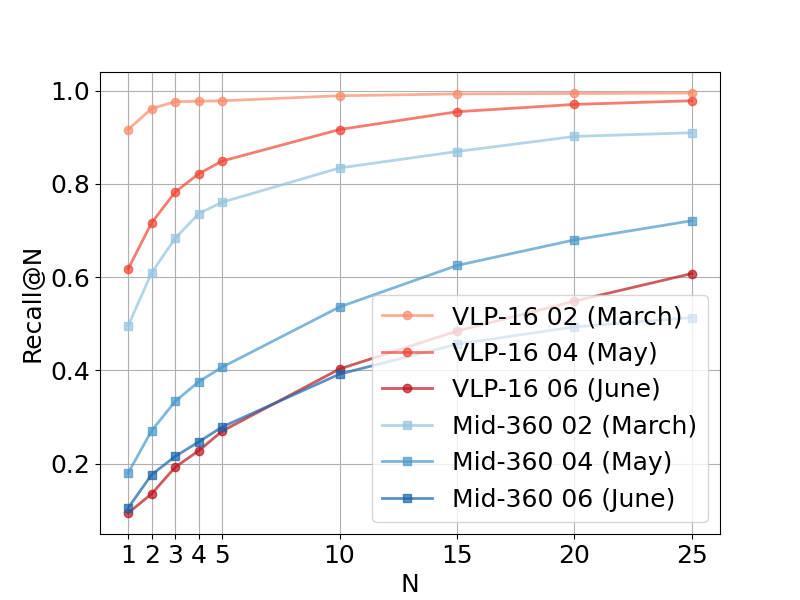}
    \caption{Place recognition: Recall graph for different values of N. Blue curves are obtained with the Livox Mid-360, the red curves with the Velodyne VLP-16 LiDAR in different seasons.}
    \label{recall_graph}
\end{figure}

\begin{figure}[ht]
    \centering
    \includegraphics[width=\linewidth]{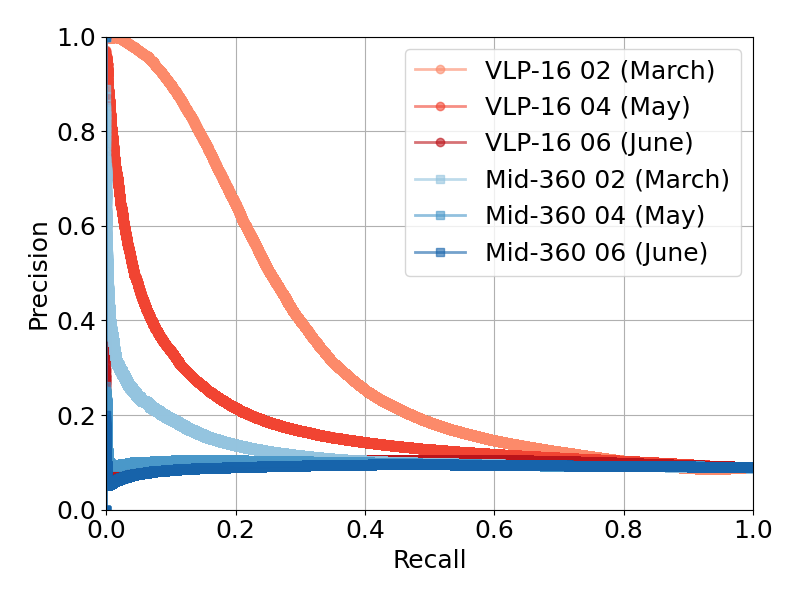}
    \caption{Place recognition: Precision-recall curve for different season generalization scenarios. Blue curves are obtained with the Livox Mid-360, the red curves with the Velodyne VLP-16 LiDAR.}
    \label{pr_curve_graph}
\end{figure}

\subsection{Place recognition results} 
To evaluate the place recognition capabilities of our dataset, we used a handcrafted global descriptor, Scan Context \cite{kim2018scan}, and two state-of-the-art deep learning global descriptors, PointNetVLAD \cite{uy2018pointnetvlad} and MinkLoc3Dv2 \cite{komorowski2022improving}. Following an inter-session setup, we tested how well the method generalizes when revisiting the same vineyard locations at different stages of the vine growth cycle. For this evaluation, we only considered the vineyard sequences, excluding the pergola. Specifically, we focused on the Run 01 recordings, which provide the most extensive coverage of the vineyard terrain. To highlight the impact of seasonal changes on agricultural place recognition, we used the first run (recorded in February) as the database, while the subsequent monthly campaigns were used as queries in each iteration. Table \ref{pr-results-table-sc} reports the Scan Context quantitative results for all campaigns, using Recall@5 and Precision@5 under different positive match thresholds. On the other hand, Table \ref{pr-results-table-dl} shows the results for the selected months with the two learned descriptors in terms of Recall@1\% and Recall@1, using Run 01 from the selected campaign as the training set and Run 02 as the evaluation set. A correct match is selected in a maximum range of 5 meters. 

The results show high performance when queries belong to the same season database, but performance decreases as the crop undergoes significant changes in subsequent months. Furthermore, the geometry-based descriptor Scan Context outperforms the two learned methods, which were designed for urban environments. This evidences the challenges posed by agricultural environments for localization. Figures \ref{recall_graph} and \ref{pr_curve_graph} present the results in graphical form, showing Recall@N values and the precision–recall curves across the different months.

\section{CONCLUSIONS}
The TEMPO-VINE dataset is a multi-temporal dataset for benchmarking sensor fusion, SLAM, and place recognition research methods in a realistic vineyard environment. Robotics and autonomous driving solutions are often tested in simulated or isolated field experiments, often lacking the necessary validation process to introduce robust and continuous operations on the field. Vineyards are a strongly dynamic and challenging environment where automation solutions are raising interest of both research and industry. TEMPO-VINE is the first public dataset comprising an extensive data collection on large-scale vineyards with multiple rows longer than $100$ m. The benchmark is fully ROS-compatible and it offers raw data and ground truth trajectories for multiple runs in a standard trellis and also in a pergola architecture, which was missing in the landscape of agricultural datasets. Nonetheless, TEMPO-VINE is a heterogeneous LiDAR dataset, mixing advanced 3D LiDAR such as the Velodyne VLP-16 with the more recent and low-cost sensing solution offered by the Livox Mid-360. Platform cost reduction is another fundamental research goal that must be explored to lead automation to worldwide success and diffusion, especially for small agricultural realities.

Through benchmark evaluations of both SLAM and place recognition tasks, we demonstrated the limitations in reaching robust performance from winter to summer, underscoring the need for further sensor fusion and SLAM research in these real settings.

As future work, the dataset will be extended to reach two years temporal coverage, including all the four seasons, and different trajectories. SLAM benchmarking will be enriched and open to the community, carrying out introspection analyses about strengths and failures of methods over the different field conditions. By providing multi-seasonal, multi-modal data under different vineyards and environmental conditions, TEMPO-VINE raises the bar of autonomous navigation benchmarking in agriculture, paving the way for the next generation of robotics and automation solutions.



\bibliographystyle{IEEEtran}
\bibliography{bibliography}

\end{document}